\def\year{2019}\relax
\documentclass[letterpaper]{article} 
\usepackage{aaai19}  
\usepackage{times}  
\usepackage{helvet}  
\usepackage{courier}  
\usepackage{url}  
\usepackage{graphicx}  

\usepackage[justification=raggedright]{caption}	
\usepackage[lined,ruled,linesnumbered]{algorithm2e}
\usepackage[T1]{fontenc}    
\usepackage[utf8]{inputenc} 
\usepackage{amsfonts}       
\usepackage{amssymb}
\usepackage{amsmath}
\usepackage{amsthm}
\usepackage{bbm}
\usepackage{bm}
\usepackage{booktabs}       
\usepackage{caption}
\usepackage{color}
\usepackage{commath}
\usepackage{comment}
\usepackage{epsfig}
\usepackage{float}
\usepackage{hyperref}       
\usepackage{lscape}
\usepackage{mathtools}
\usepackage{microtype}      
\usepackage{multirow}
\usepackage{nicefrac}       
\usepackage{paralist}
\usepackage{soul}
\usepackage{subcaption}
\usepackage{textcomp}
\usepackage{xcolor}
\usepackage{blindtext}
\usepackage{comment}

\def\etal{et~al.~}			  
\def\eg{e.g.,~}                

\newcommand{\X}{\mathbf{X}}
\newcommand{\x}{\mathbf{x}}
\newcommand{\Y}{Y}
\newcommand{\y}{y}

\newcommand{\E}{\mathbb{E}}
\newcommand{\I}{\mathbbm{1}}
\newcommand{\sign}{\textrm{sign}}

\newcommand{\Loss}{\mathbb{L}}

\newcommand{\RD}{\mathbb{RD}}
\newcommand{\crd}{C^{\eta}}

\newlength\paramargin
\newlength\figmargin
\newlength\secmargin

\usepackage{color}
\long\def\ignorethis#1{}

\newtheorem{theorem}{Theorem}

\newtheorem{corollary}[theorem]{Corollary}

\newtheorem{formulation}{Problem Formulation}
\newtheorem{definition}{Definition}

\begin{document}
%
\title{Fairness-aware Classification: Criterion, Convexity, and Bounds}
\author{Lu Zhang, Yongkai Wu, and Xintao Wu\\
University of Arkansas\\
\{lz006,yw009,xintaowu\}@uark.edu}

\maketitle
\begin{abstract}
Fairness-aware classification is receiving increasing attention in the machine learning fields. Recently research proposes to formulate the fairness-aware classification as constrained optimization problems. However, several limitations exist in previous works due to the lack of a theoretical framework for guiding the formulation. In this paper, we propose a general framework for learning fair classifiers which addresses previous limitations. The framework formulates various commonly-used fairness metrics as convex constraints that can be directly incorporated into classic classification models. Within the framework, we propose a constraint-free criterion on the training data which ensures that any classifier learned from the data is fair. We also derive the constraints which ensure that the real fairness metric is satisfied when surrogate functions are used to achieve convexity. Our framework can be used to for formulating fairness-aware classification with fairness guarantee and computational efficiency. The experiments using real-world datasets demonstrate our theoretical results and show the effectiveness of proposed framework and methods.
\end{abstract}

\section{Introduction}
Fairness-aware classification is receiving increasing attention in the machine learning fields. Since the classification models seek to maximize the predictive accuracy, some individuals may get unwanted digital bias when the models are deployed for making predictions. As fairness becomes a more and more important requirement in machine learning, it is imperative to ensure that the learned classification models can strike a balance between accurate predictions and fair predictions. Previous works on this topic can be mainly categorized into two groups: the in-processing methods which incorporate the fairness constraints into the classic classification models (e.g., \cite{Kamishima2011a,Goh,Krishna2018,Zafar2017b,Zafar2017a}), and the pre/post-processing methods which modify the training data and/or derive fair predictions based on the potentially unfair predictions made by the classifier (e.g., \cite{Feldman2015,Hardt2016,Zhang2017a,Zhang2017c,zhang2018achieving}). In this work, we focus on the in-processing methods.

Very recently, several works have been proposed for formulating the fairness-aware classification as constrained optimization problems \cite{Goh,Zafar2017b,Zafar2017a}. Generally, they aim to minimize a loss function subject to certain fairness constraints. Although the idea is reasonable and rather straightforward, there still exist a number of challenges. As a result, several limitations exist in the previous works. One challenge is how to formulate the classic fairness notions (such as demographic parity) as convex constraints in the optimization, which is not well-addressed in previous works. In \cite{Zafar2017a}, the authors propose the decision boundary fairness, which is a linear constraint for margin-based classifiers. However, the authors fail to explicitly show the connection between the decision boundary fairness and classic fairness notions, where the latter is the one people really care about. Another work \cite{Zafar2017b} suffers a similar issue. In addition, its formulated optimization problem is non-convex and difficult to solve efficiently. In \cite{Goh}, a convex constraint is derived from the risk ratio, a classic fairness measure. However, the constraint is enforced from one direction only, i.e., it cannot avoid reverse bias. Another challenge is that, when surrogate functions are used to convert non-convex functions to convex functions, which is a widely-used strategy to achieve convexity in optimization and adopted in many related works, estimation errors must exist due to the difference between the surrogate function and the original non-convex function. Thus, achieving the constraints represented by surrogate functions does not necessarily mean achieving the real fairness criterion. To the best of our knowledge, no work has considered this gap produced by the estimation errors due to the use of the surrogate function.

In this paper, we propose a general framework for fairness-aware classification which addresses all above issues. The framework formulates various commonly-used fairness metrics (risk difference, risk ratio, equal odds, etc.) as convex constraints that are then directly incorporated into classic classification models. Within the framework, we present a constraint-free criterion on the training data which ensures that any classifier learned from the data will be fair. Thus, when the criterion is satisfied, there is no need to add any fairness constraint into optimization for learning fair classifiers. When the criterion is not satisfied, we need to learn fair classifiers by solving the constrained optimization problems. To connect the surrogate function represented fairness constraints to the real fairness metric, we further derive the lower and upper bounds of the real fairness measure based on the surrogate function, and develop the refined fairness constraints. This means that, if the refined constraints are satisfied, then it is guaranteed that the real fairness measure is also bounded within the given interval. The bounds work for any surrogate function that is convex and differentiable at zero with the derivative larger than zero.
In the experiments, we evaluate our method and compare with previous works using the real-world datasets. The results demonstrate the correctness of the constraint-free criterion. For learning fair classifiers, the results show that our method achieves better fairness performance than previous methods. In addition, for the same fairness performance level, our method also consistently outperforms previous methods in terms of the predictive accuracy.


\section{The Fairness-aware Classification Framework}\label{sec:fcf}
In this section we present our fairness-aware classification framework. We first introduce the unconstrained optimization formulation for the classic classification models as proposed in \cite{Bartlett2006}, and then present our constrained optimization formulation for fairness-aware classification. Throughout the paper, we use the vector of variables $\X \in \mathcal{X}$ to denote the features used in classification, and the binary variable $\Y \in \mathcal{Y} = \{-1,1\}$ to denote the binary label. The training data $\mathbb{D} = \{ (\x_i, \y_i)\}_{i=1}^{N}$ is a sample drawn from an unknown but fixed distribution.

\subsection{Classification Problem}
The learning goal of classification is to find a classifier: $f \colon \mathcal{X} \mapsto \mathcal{Y}$ that minimizes the average of the classification loss (a.k.a the empirical loss):
\begin{equation*}
	\Loss (f) =  \E_{\X, \Y} [ \I_{f(\x) \neq \y} ] ,
\end{equation*}
where $\I_{[\cdot]}$ is an indicator function and we let $ \I_{\mathrm{true}} = 1, \I_{\mathrm{false}}=0 $. Thus, the classification problem can be formulated as an optimization problem:
\begin{align*}
	\min_{f \in \mathcal{F}} \Loss(f) = \min_{f \in \mathcal{F}} \E_{\X, \Y} [ \I_{f(\x) \neq \y} ].
\end{align*}
Directly solving this optimization problem is intractable since the objective function is non-convex \cite{Bartlett2006}. For efficient computation, another predictive function $h$ is adopted which is performed in real number domain $\mathcal{R}$, i.e., $h \colon  \mathcal{X} \mapsto \mathcal{R}$, and let $f = \sign(h) $ once $h$ is learned. Thus, the empirical loss is reformulated as
\begin{align*}
&	\Loss (f) = \E_{\X, \Y} \big[ \I_{\sign \big( h(\x) \big) \neq \y} \big]                     \\
&	          = \E_{\X} \big[ Pr(\Y=1|\x) \I_{h(\x) < 0} +    Pr(\Y=-1|\x) \I_{h(\x) > 0} \big].
\end{align*}
Then, the indicator function (a.k.a 0-1 loss function) is replaced with a convex surrogate function $\phi(\cdot)$. As a result, the empirical loss is written as
\begin{align*}
	\Loss_{\phi} (h) & = \E_{\X} \big[ Pr(Y=1|\x) \phi \big(h(\x) \big)            \\
	                 & + \big(1 - Pr(Y=1|\x) \big) \phi \big( - h(\x) \big) \big],
\end{align*}
which is also known as the $\phi$-loss, and the optimization problem is reformulated as
\begin{eqnarray}
	\min_{ h \in \mathcal{H}} \Loss_{\phi} (h).
	\label{opt:standard_classification}
\end{eqnarray}
In the past decades, a number of surrogate loss functions have been proposed and well studied, such as the hinges loss, the square loss, the logistic loss, and the exponential loss.

\subsection{Fairness-aware Classification}
The fairness-aware classification aims to find a classifier that minimizes the empirical loss while satisfying certain fairness constraints. Several fairness notions or definitions are proposed in the literature, such as demographic parity \cite{Pedreshi2008}, mistreatment parity \cite{Zafar2017b}, calibration \cite{Pleiss2017a}, etc., and research shows that different fairness notions are generally incomparable with each other and cannot be satisfied simultaneously \cite{Kleinberg2016}. Our framework is not limited to a specific fairness notion. In this paper, we present our framework based on the demographic parity. In the appendix, we show how the framework can be easily generalized to other fairness notions.

Demographic parity is the most widely-used fairness notion in the fairness-aware learning field. It requires the decision made by the classifier is independent to certain sensitive attribute, such as sex or race. We denote the sensitive attribute by $S$, assuming that it is associated with two values: sensitive group $s^{-}$ and non-sensitive group $s^{+}$. Usually, demographic parity is quantified with regard to risk difference, i.e., the difference of the positive predictions between the sensitive group and non-sensitive group. For example, in hiring, risk difference can be given by the probability difference of being classified as hired between male applicants and female applicants. Using the same language as that in the previous subsection, the risk difference produced by a classifier $f$ is expressed as
\begin{align*}
	\RD (f) = \E_{\X|S=s^+} [\I_{f(\x) = 1}] - \E_{\X|S=s^-} [\I_{f(\x) = 1}]. &
\end{align*}
As a quantitative metric, we say that classifier $f$ is considered as fair if $ |\RD (f)| \leq \tau$, where $ \tau $ is the user-defined threshold. For instance,
the 1975 British legislation for sex discrimination sets $\tau = 0.05$.

By directly incorporating the risk difference into the optimization problem, we obtain
\begin{align}
	\min_{ h \in \mathcal{H}} \quad & \Loss_\phi (h)     \label{opt:constained_fair_classification} \\
	\text{subject to }        \quad & \RD (f) \leq \tau, \quad  -\RD (f) \leq \tau. \nonumber
\end{align}
Obviously, the above optimization problem is non-convex. Similar to the loss function, we adopt surrogate functions to convert the risk difference to convex constraints. By using predictive function $h$ and the indicator function, we can rewrite the risk difference as

\begin{align*}
	\RD (f) & = \E_{\X|S=s^+} \Big[ \I \big[ \sign \big( h(\x) \big) = 1 \big] \Big]                             \\
	& - \E_{\X|S=s^-} \Big[ \I \big[ \sign \big( h(\x) \big) = 1 \big] \Big]                            \\
	 & = \E_{\X|S=s^+} [\I_{h(\x) >0}] - \E_{\X|S=s^-} [\I_{h(\x) > 0}]       \\
	 & = \E_{\X|S=s^+} [\I_{h(\x) >0}] + \E_{\X|S=s^-} [\I_{h(\x) < 0}]  - 1.
\end{align*}

\begin{figure}
	\centering
	\includegraphics[width=0.8\linewidth]{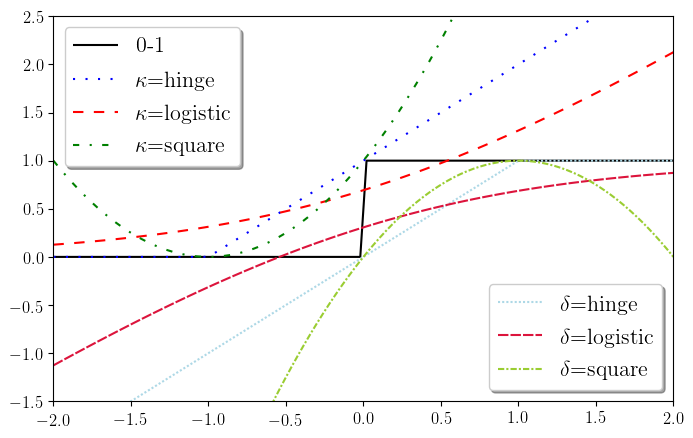}
	\captionof{figure}{Examples of $\kappa(\cdot)$ and $\delta(\cdot)$. }
	\label{fig:surrogate}
\end{figure}

It follows that
\begin{align}
	\label{eq:rd1}
	\RD(f) =  & \E_{\X} \big[ \frac{P(S=s^+|\x)}{P(S=s^+)} \I_{h(\x)>0} \\
	\nonumber & + \frac{P(S=s^-|\x)}{P(S=s^-)} \I_{h(\x)<0} -1 \big].
\end{align}
For simplicity, we may want to denote $P(S=s^+|\x)$ by $\eta(\x)$ and $P(S=s^+)$ by $p$.
Similarly, the indicator function in above equation can be replaced with the surrogate function. The issue here is, two constraints $\RD (f) \leq \tau$ and $-\RD (f) \leq \tau$ are opposite to each other. Thus, replacing all indicator functions with a single surrogate function will result in a convex-concave problem, where only heuristic solutions for finding local optima are known to exist. Therefore, we adopt two surrogate functions, a convex one $\kappa(\cdot)$ and a concave one $\delta(\cdot)$, each of which replaces the indicator function for one constraint.
As a result, the formulated constrained optimization problem is convex and can be efficiently solved. We call the risk difference represented by $\kappa(\cdot)$ and $\delta(\cdot)$ as the $\kappa,\delta$-risk difference, denoted by $\RD_{\kappa}(h)$ and $\RD_{\delta}(h)$.
Almost all commonly-used surrogate functions can be adopted for $\kappa(\cdot)$ and $\delta(\cdot)$, by performing some shift or flip.
Examples of $\kappa(\cdot)$ and $\delta(\cdot)$ are shown in Figure~\ref{fig:surrogate}.

\begin{table}[ht]
	\centering
	\begin{tabular}{|c|cc|}
		\hline
		\multirow{2}{*}{\texttt{Sex}} & \multicolumn{2}{c|}{\texttt{GPA}}                \\
		                              & \texttt{High}                     & \texttt{Low} \\ \hline
		\texttt{Male}                 & 51                                & 49           \\
		\texttt{Female}               & 48                                & 52           \\ \hline
	\end{tabular}
	\caption{An example of admitting students.}
	\label{tab:demo1}
\end{table}

To sum up, we obtain the following convex optimization formulation for learning fair classifiers.

\begin{formulation}\label{for:fc1}
	The goal of the fairness-aware classification is to find a classifier $f$ which minimizes the empirical loss $\Loss(f)$ while satisfying fairness constraint $|\RD(f)|\leq \tau$. It can be approached by solving the following constrained optimization problem
	\begin{align}
		\label{opt:surrogated_fair_classification} \min_{ h \in \mathcal{H}} \quad & \Loss_{\phi} (h)      \nonumber                                  \\
		\nonumber                                   \text{ subject to } \quad      & \RD_{ \kappa } (h) \leq c_1, \quad -\RD_{ \delta } (h) \leq c_2,
	\end{align}
	where $f = \sign(h)$, $\kappa(\cdot)$ is a convex surrogate function, $\delta(\cdot)$ is a concave surrogate function, $\eta(\x) = P(S=s^+|\x)$, $p = P(S=s^+)$, $c_1, c_2$ are the thresholds of the $\kappa,\delta$-risk difference, and
	\begin{align*}
		 & \Loss_{\phi} (h) = \E_{\X} \big[ Pr(Y=1|\x) \phi \big(h(\x) \big)                                                                           \\
		 & \qquad\qquad + \big(1 - Pr(Y=1|\x) \big) \phi \big( - h(\x) \big) \big],                                                                          \\
		 & \RD_{ \kappa } (h) = \E_{\X} \big[ \frac{\eta(\x)}{p} \kappa \big( h(\x) \big) + \frac{1-\eta(\x)}{1-p} \kappa \big( -h(\x) \big) -1 \big], \\
		 & \RD_{ \delta } (h) = \E_{\X} \big[ \frac{\eta(\x)}{p} \delta \big( h(\x) \big) + \frac{1-\eta(\x)}{1-p} \delta \big( -h(\x) \big) -1 \big].
	\end{align*}
\end{formulation}

Next, we will present two important results within the framework, namely a constraint-free criterion on the training data that ensures fairness for any classifier learned from it, as well as the lower and upper bounds of the risk difference using the $\kappa,\delta$-risk difference.

\section{The Constraint-free Criterion}\label{sec:cfc}
Adding constraints into the classification models increases the computational complexity and also decreases the predictive accuracy. It is desired not to incorporate any fairness constraint if it is guaranteed that the classifier learned will be fair. This situation is possible. Consider an example of admitting students. The application profile contains two attributes, a sensitive attribute \texttt{Sex} and a non-sensitive attribute \texttt{GPA}. The statistics of the dataset is shown in Table \ref{tab:demo1}. Assume that classifiers used for making decisions are based on \texttt{GPA} (as the use of sensitive attributes is usually prohibited), then there are a total of four possible deterministic classifiers: accepting all students, accepting all the students with \texttt{GPA = High}, accepting all the students whose \texttt{GPA = Low}, and accepting none.  The corresponding risk differences of the four classifiers are 0, 0.02, -0.03, and 0 respectively, which are all considered to be fair based on a $\tau=0.05$ threshold. In this case, no matter which classifier is learned, the predictions will always be fair.


In this section, we propose a constraint-free criterion of ensuring fairness. We first define two special classifiers, and then show in Theorem \ref{thm:mrdc} that they provide the maximum and minimum of risk difference that any classifier can have. Here we skip the proof of Theorem \ref{thm:mrdc} which is included in the appendix.

\begin{definition}
	The maximal risk difference classifier $f_{\max}$  and the minimal risk difference classifier $f_{\min}$ are defined as:
	\begin{equation*}
		f_{\max} (\x)=
		\begin{cases}
			\phantom{-}1 & \text{ if } \eta (\x) \geq p , \\
			-1           & \text{ otherwise, }
		\end{cases}
	\end{equation*}
	\begin{equation*}
		f_{\min} (\x)=
		\begin{cases}
			-1           & \text{ if } \eta (\x) \geq p , \\
			\phantom{-}1 & \text{ otherwise.}
		\end{cases}
	\end{equation*}
	\label{def:mrdc}
\end{definition}

\begin{theorem}
	For any classifier $f \colon \mathcal{X} \mapsto \mathcal{Y}$, it always holds that $\RD^{-} \leq \RD (f) \leq \RD^{+}$, where $\RD^{-} = \RD ( f_{\min} )$ and $\RD^{+} = \RD ( f_{\max} )$.
	\label{thm:mrdc}
\end{theorem}

From Theorem \ref{thm:mrdc}, we directly obtain Corollary \ref{cor:2}.

\begin{corollary}\label{cor:2}
	Given threshold $\tau$, for a training data if we have $\RD^{+}  \leq \tau$ and $\RD^{-}  \geq -\tau$, then any classifier learned from this dataset is fair.
\end{corollary}

\section{Bounding Fairness Constraints with Surrogate Functions}\label{sec:b}
\begin{figure*}
	\centering
	\begin{subfigure}{0.45\textwidth}
		\centering
		\includegraphics[width=0.7\textwidth]{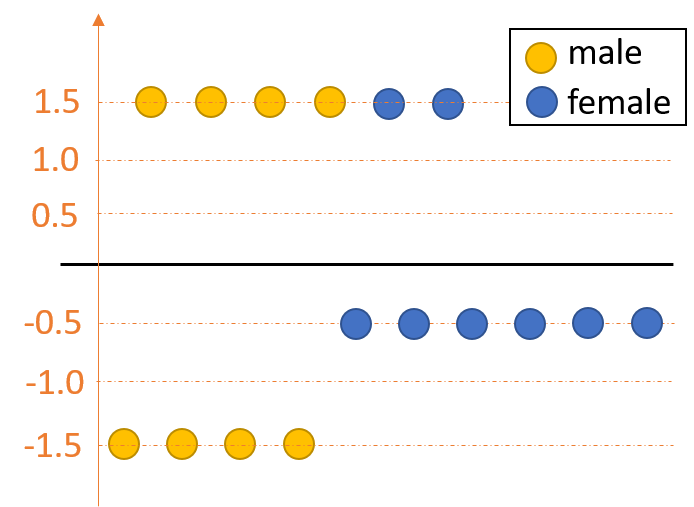}
		\caption{A classifier that meets the $\kappa,\delta$-risk difference constraint makes unfair predictions.}
		\label{fig:demo1}
	\end{subfigure}%
	\vspace{1em}%
	\begin{subfigure}{0.45\textwidth}
		\centering
		\includegraphics[width=0.7\textwidth]{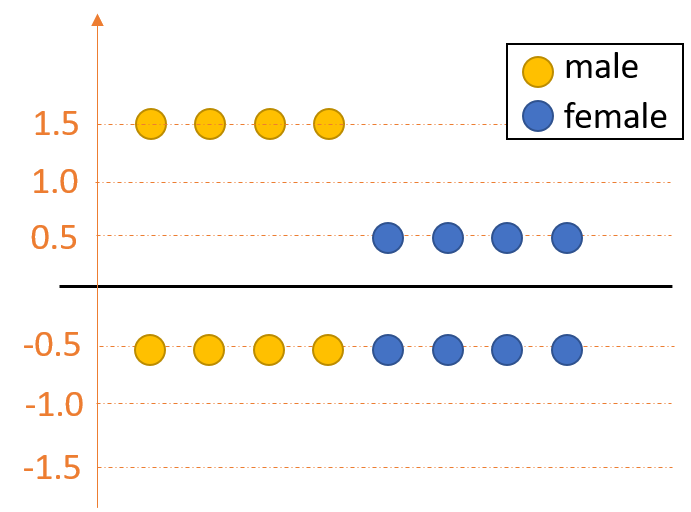}
		\caption{A classifier that doesn't meet the $\kappa,\delta$-risk difference constraint makes fair predictions.}
		\label{fig:demo2}
	\end{subfigure}
	\caption{Two classifiers and their predictions.}
	\label{fig:demo}
\end{figure*}

When the constraint-free criterion is not satisfied, the next step will be learning fair classifiers based on Problem Formulation \ref{for:fc1}. However, the use of the surrogate function will inevitably produce estimation errors. This means that satisfying constraints for the $\kappa,\delta$-risk difference, i.e., $\RD_{\kappa}(h)\leq \tau$ and $-\RD_{\delta}(h)\leq \tau$ does not mean that the constraint for the risk difference is also satisfied, i.e., $|\RD(f)|\leq \tau$. Consequently, solving the the constrained optimization problem does not necessarily result in a fair classifier based on the real risk difference. In fact, there is even no any fairness guarantee on the produced classifier. We use an example to show this. Consider two margin-based classifiers where the surrogate functions are linear functions of the margin from the data point to the decision boundary. Therefore, the risk difference is computed by counting the number of data points above and below the decision boundary, and the $\kappa,\delta$-risk difference is computed by measuring the average signed distance from the data points to the decision boundary. In the dataset shown in Figure~\ref{fig:demo1}, we obtain that the $\kappa,\delta$-risk difference is 0 but the real risk difference is 0.25. This means that a classier obtained by solving the constrained optimization problem actually can be very unfair. In the dataset shown in Figure~\ref{fig:demo2}, the risk difference is 0 but the $\kappa,\delta$-risk difference is 0.5, meaning that some fair classifiers cannot be obtained by solving the constrained optimization problem.

In this section, we present the method for deriving the lower and upper bounds for the risk difference using the $\kappa,\delta$-risk difference, which provide a fairness guarantee for our framework. The method works for various types of surrogate functions (e.g., hinge, square, logistic, exponential, etc.). We begin with defining the conditional risk difference $\crd \big( h(\x) \big)$:
\[ \crd \big( h(\x) \big) = \frac{\eta(\x)}{p} \I_{h(\x)>0} + \frac{1-\eta(\x)}{1-p} \I_{h(\x)<0} - 1. \]
Then, according to Eq. \eqref{eq:rd1}, we have $\RD(f) = \E_{\X} [ \crd \big( h(\x) \big) ]$. When surrogate function $\kappa(\cdot)$ (resp. $\delta(\cdot)$) is adopted, we similarly define the conditional $\kappa$-risk difference 
\begin{equation*}
\crd_{\kappa} \big( h(\x) \big) = \frac{\eta (\x)}{p} \kappa \big( h( \x ) \big)  + \frac{1 - \eta (\x)}{1-p} \kappa \big( - h( \x ) \big) - 1,
\end{equation*}
and we have $\RD_{\kappa} (h) = \E_\X \big[\crd_{\kappa} \big( h(\x) \big) \big]$.

Note that the values of $\crd \big( h(\x) \big)$ and $\crd_{\kappa} \big( h(\x) \big)$ depend on $\eta (\x)$ and $h( \x )$, which are determined by the subpopulation of the data specified by $\x$, as well as predictive function $h$. In order to study the general situations for any specific subpopulation and any possible predictive function, we define the generic conditional risk difference $\crd(\alpha)$ and the generic conditional $\kappa$-risk difference $\crd_{\kappa}(\alpha)$:
\[ \crd(\alpha) = \frac{\eta}{p} \I_{\alpha>0} + \frac{1-\eta}{1-p} \I_{\alpha<0} - 1, \]
\[ \crd_{\kappa} (\alpha) = \frac{ \eta }{p} \kappa ( \alpha)  + \frac{1-\eta}{1-p} \kappa (-\alpha ) - 1, \]
for any $\eta\in [0,1]$ and $\alpha\in \mathcal{R}$. Then, the minimal conditional risk difference $H^{-} (\eta)$ and the minimal conditional $\kappa$-risk difference $H^{-}_{\kappa} (\eta)$ for any specific subpopulation and any possible predictive function are given by
\begin{equation*}
	H^{-} (\eta) = \min_{\alpha \in R} \crd (\alpha) = \min_{\alpha \in R} \big[ \frac{\eta}{p} \I_{\alpha>0} + \frac{ 1- \eta}{1-p} \I_{\alpha<0} -1 \big],
\end{equation*}
\begin{equation}
	\label{eq:H:inf} H^{-}_{\kappa} (\eta) = \min_{\alpha \in R} \crd_\kappa (\alpha) = \min_{\alpha \in R} \big[ \frac{\eta}{p} \kappa ( \alpha)  + \frac{ 1- \eta}{1-p} \kappa ( - \alpha ) -1   \big].
\end{equation}
As a result, it is straightforward that the minimal risk difference achieved by any predictive function (i.e., $\RD^{-}$) is also the expectation of $H^{-} (\eta(\x))$ since for any possible input $\x$ $H^{-} (\eta(\x))$ provides the minimal conditional risk difference.
Similarly, the minimal $\kappa$-risk difference achieved by any predictive function (denoted by $\RD^{-}_{\kappa}$) is the expectation of $H^{-}_{\kappa} (\eta(\x))$, as given by 
\[ \RD^{-}_{\kappa} = \E_\X \big[ H^{-}_{\kappa} (\eta(\x)) \big]. \]
Finally, we define the minimal conditional $\kappa$-risk difference within interval $ \alpha \text{ s.t. } \alpha( \eta_S - p ) \geq 0 $:
\begin{eqnarray}
	H^{\circ}_{\kappa} (\eta) = \min_{\alpha : \alpha( \eta - p ) \geq 0} \crd_\kappa (\alpha).
	\label{eq:H:secondary}
\end{eqnarray}

We similarly define $H^{+}(\eta)$ the maximal conditional risk difference, $H^{+}_{\delta} (\eta)$ the maximal conditional $\delta$-risk difference, $\RD^{+}_{\delta}$ the maximal $\delta$-risk difference, as well as $H^{\circ}_{\delta} (\eta)$ the minimal conditional $\delta$-risk difference within interval $ \alpha \text{ s.t. } \alpha( \eta_S - p ) \geq 0 $.

Now, we are able to present our results, which are given in Theorem \ref{thm:psi} and Corollary \ref{thm:bounds}. The proof is skipped here and can be found in the appendix.

\begin{theorem}
	\label{thm:psi}
	If $\kappa(\cdot)$ is convex and differentiable at zero with $\kappa^{\prime}(0) > 0 $, $\delta(\cdot)$ is concave and differentiable at zero with $\delta^{\prime}(0) > 0 $, then for any predictive function $h$, we have
	\begin{eqnarray*}
		\label{eq:fair_calibrated}
		\psi_{\kappa} ( \RD(h) - \RD^{-} ) \leq \RD_{\kappa}(h) - \RD_{\kappa}^{-}, \\
		\psi_{\delta} ( \RD^{+} - \RD(h) ) \leq \RD_{\kappa}^{+} - \RD_{\kappa}(h),
	\end{eqnarray*}
	where
	\begin{eqnarray*}
		\label{eq:psi1}
		\psi_{\kappa}(\mu) = H^{\circ}_{\kappa} \big( p ( 1- p) \mu + p ) - H^{-}_{\kappa} \big( p ( 1- p) \mu + p \big) , \\
		\label{eq:psi2}
		\psi_{\delta}(\mu) = H^{+}_{\delta} \big( p ( 1- p) \mu + p \big) - H^{\circ}_{\delta} \big( p ( 1- p) \mu + p ) .
	\end{eqnarray*}
\end{theorem}

\begin{corollary}\label{thm:bounds}
	For any predictive function $h$, let classifier $f = \sign(h)$, if $\kappa(\cdot)$ is convex and differentiable at zero with $\kappa^{\prime}(0) > 0 $, $\delta(\cdot)$ is concave and differentiable at zero with $\delta^{\prime}(0) > 0 $, and $H^{-}_{\kappa} (\eta) < H^{\circ}_{\kappa} (\eta)$ and $H^{+}_{\delta} (\eta) > H^{\circ}_{\delta} (\eta) $ hold for all $\eta \in [0, 1], \eta \neq p$,  then risk difference $\RD(f)$ is bounded by following inequalities\footnote{Based on Scott \etal \cite{scott2012}, $\psi_{\kappa}(\cdot)$ and $\psi_{\delta}(\cdot)$ are invertible if $H^{\circ}_{\kappa} (\eta) > H^{-}_{\kappa} (\eta)$ and $H^{\circ}_{\delta} (\eta) < H^{+}_{\delta} (\eta)$ for all $\eta \in [0, 1], \eta \neq p$.}:
	\begin{eqnarray*}
		\RD (f) \leq \RD^{-} + {\psi_{\kappa}}^{-1} \big( \RD_{ \kappa } (h) - \RD^{-}_{ \kappa } \big), \\
		\RD (f) \geq \RD^{+} - {\psi_{\delta}}^{-1} \big( \RD^{+}_{ \delta } - \RD_{ \delta } (h) \big).
	\end{eqnarray*}
\end{corollary}

Based on the upper and lower bounds of $\RD(f)$, we modify Problem Formulation \ref{for:fc1} to obtain Problem Formulation \ref{for:fc2} with refined fairness constraints which guarantee the real fairness requirement.

\begin{formulation}\label{for:fc2}
	A fair classifier $f = \sign(h)$ that achieves fairness constraint $-c_{2}\leq \RD(f)\leq c_{1}$ can be obtained by solving the following constrained optimization
	\begin{align}
		\label{opt:refined_fair_classification}
		\min_{ h \in \mathcal{H}} \quad                                   & \Loss_{\phi} (h)                                                               \\
		\nonumber                               \text{ subject to } \quad & \RD_{ \kappa } (h) \leq \psi_{\kappa}( c_{1} - \RD^{-} ) + \RD^{-}_{ \kappa },  \\
		\nonumber                                                         & -\RD_{ \delta } (h) \leq \psi_{\delta}( c_{2} + \RD^{+} ) + \RD^{+}_{ \delta }.
	\end{align}
\end{formulation}
Note that the RHS of above two inequalities are constants for a given dataset. Therefore, the constrained optimization problem is still convex. As stated, we can adopt almost any type of surrogate function for $\kappa(\cdot)$ and $\delta(\cdot)$. Some commonly-used surrogate functions are listed in Tables \ref{tab:surrogate_functions}. Their corresponding $\psi_{\kappa}(\cdot)$ and $\psi_{\delta}(\cdot)$ are derived and shown in the last column where the derivation details are skipped.

\begin{table*}[ht]
	\centering
	\renewcommand{\arraystretch}{1.6}
	\begin{tabular}{|l|c|c|c|}
		\hline
		Name of $\kappa\text{-}\delta$ & $\kappa( \alpha)$ for $\alpha \in \textbf{R}$ & $\delta( \alpha)$ for $\alpha \in \textbf{R}$ & $\psi_{\kappa} (\mu)$ or $\psi_{\delta} (\mu)$ for $\mu \in (0, 1/p]$ \\ \hline
		Hinge                          & $\max \{ \alpha + 1, 0 \}$                    & $\min \{ \alpha, 1 \}$                        & $\mu $                                                              \\
		Square                         & $(\alpha + 1 )^2 $                            & $1 - (1 - \alpha )^2 $                        & $\mu^2$                                                             \\
		Exponential                    & $\exp(\alpha) $                               & $1 - \exp(-\alpha) $                          & $(\sqrt{(1-p)\mu+1} - \sqrt{1-p \mu} )^2$                                             \\ \hline
	\end{tabular}
	\caption{Some common surrogate functions for $\kappa \text{-} \delta$ and the corresponding $\psi_{\kappa} (\mu)$ and $\psi_{\delta} (\mu)$.}
	\label{tab:surrogate_functions}
\end{table*}

\section{Experiments}\label{sec:e}
\subsection{Experimental Setup}
{\bf Dataset.} In the experiments we use two datasets: Adult and Dutch. The Adult dataset \cite{Lichman2013} contains a total of 48,842 instances, each of which is characterized by 14 attributes (\eg \texttt{sex}, \texttt{age}, \texttt{work\_class}, \texttt{education}, \texttt{income}, etc.). We consider $\texttt{sex}$ as the sensitive attribute with two values, $\texttt{male}$ and $\texttt{female}$. Then, we binarize \texttt{income} and use it as the class label, i.e., $Y=1$ if an individual's income is above \$50K and $Y=-1$ if it is below \$50k. The Dutch dataset \cite{Zliobaite2011} contains a total of 60,420 instances, each of which is characterized by 12 attributes. Similarly, we use \texttt{sex} as the sensitive attribute, and binarize \texttt{occupation} into the high-income group and the low-income group which are used as the class label.

	{\bf Baseline.} We compare our method with two related works, referred to as Zafar-1 \cite{Zafar2017a} and Zafar-2 \cite{Zafar2017b}, both of which formulate the fairness-aware classification problem as constrained optimization problems. In \cite{Zafar2017a}, the authors quantify fairness using the covariance between the users' sensitive attribute and the signed distance between the feature vectors and the decision boundary. The fairness constraint is formulated as covariance $\leq mc$, where $c$ is the measured fairness of the unconstrained optimal classifier and $m$ is a multiplication factor $\in [0,1]$. In \cite{Zafar2017b}, the fairness is quantified similarly with the distance function being replaced with a convex non-linear function. As a result, the obtained problem is a convex-concave optimization problem. In the experiments, we adopt the Disciplined Convex-Concave Programming (DCCP) \cite{Shen2016} as proposed in \cite{Zafar2017b} for solving the convex-concave optimization problem. For our method and Zafar-1, the convex optimization problem is solved using CVXPY \cite{StevenDiamond2016}.

\subsection{Constraint-free Criterion of Ensuring Fairness}
To demonstrate the sufficiency criterion of learning fair classifiers, we build the maximal/minimal risk difference classifiers $f_{\min},f_{\max}$ for both Adult and Dutch datasets, and measure the risk differences they produce, i.e., $\RD^{-},\RD^{+}$. The results are shown in the first two rows in Table \ref{tab:exp1}. As can be seen, in both datasets we have large maximal and minimal risk differences. In order to evaluate a situation with small a risk difference, we also create a variant of Adult, referred to as Adult*, where all attributes are binarized and the sensitive attribute \texttt{sex} is shuffled to incur a small risk difference. Then, we build a number of classifiers including Linear Regression (LR), Support Vector Machine (SVM) with linear kernel, Decision Tree (DT), and Naive Bayes (NB), using the three datasets as the training data with with 5-fold cross-validation. After that, their risk differences are quantified on the testing data, as shown in the last four rows in Table \ref{tab:exp1}. We can see that all values are within $\RD^{-},\RD^{+}$ which are consistent with our criterion.

\begin{table}[]
	\centering
	\begin{tabular}{|l|r|r|r|}
		\hline
		$\RD(\cdot)$ & Adult  & Dutch  & Adult* \\ \hline
		$\RD^{+}$    & 0.967  & 0.516  & 0.046  \\
		$\RD^{-}$    & -0.967 & -0.516 & -0.046 \\ \hline
		LR           & 0.371  & 0.185  & 0.000  \\
		SVM          & 0.434  & 0.156  & 0.001  \\
		DT           & 0.316  & 0.184  & 0.001  \\
		NB           & 0.447  & 0.144  & 0.001  \\ \hline
	\end{tabular}
	\caption{$\RD^{+},\RD^{-}$ and risk differences of Linear Regression (LR), Support Vector Machine (SVM), Decision Tree (DT), and Naive Bayes (NB).}
	\label{tab:exp1}
\end{table}

\begin{figure*}
	\centering
	\begin{subfigure}[b]{0.4\textwidth}
		\includegraphics[width=\linewidth, height=2.3in]{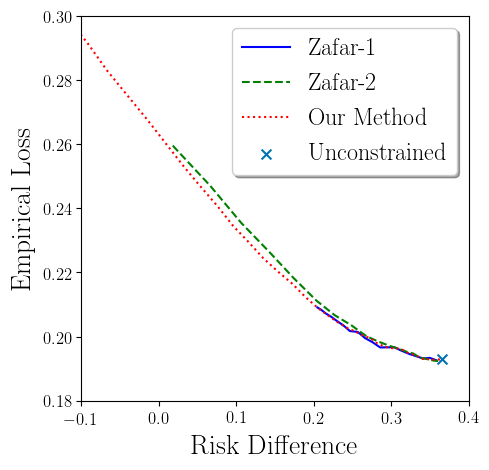}
		\caption{Adult}
		\label{fig:exp2-1}
	\end{subfigure}%
	\vspace{1em} %
	\begin{subfigure}[b]{0.4\textwidth}
		\includegraphics[width=1\linewidth, height=2.3in]{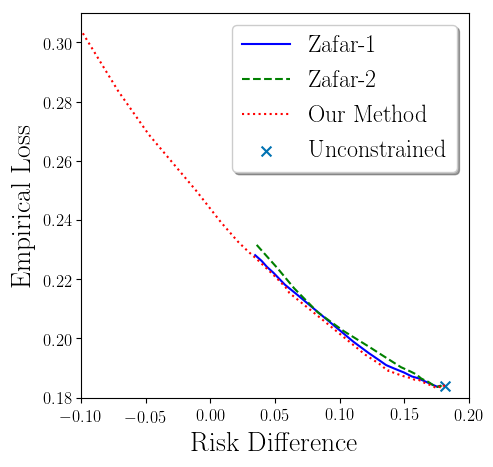}
		\caption{Dutch}
		\label{fig:exp2-2}
	\end{subfigure}
	\caption{Comparison of fair classifiers.}
	\label{fig:exp2}
\end{figure*}

\subsection{Learning Fair Classifiers}
We build our fair classifiers on both Adult and Dutch datasets by solving the optimization problem defined in Problem Formulation \ref{for:fc2}. For surrogate functions, we use the logistic function for $\phi(\cdot)$, and the hinge function for $\kappa(\cdot)$ and $\delta(\cdot)$. We also compare our methods with Zafar-1 and Zafar-2. The results are shown in Figure \ref{fig:exp2}, which depict the relationship between the obtained risk difference and empirical loss. For our method, different risk differences are obtained by adjusting relax terms $c_{1}$ and $c_{2}$, while for Zafar-1 and Zafar-2 different risk differences are obtained by adjusting the multiplication factor $m$. As can be seen, our method can achieve much smaller risk difference than Zafar-1 and Zafar-2. This may be because Zafar-1 linear functions to formulate the fairness constraints, which may incur large estimation errors; while Zafar-2 formulates a convex-concave optimization problem, where only local optima can be reached. For the same reason, we can observe that our method produces better empirical loss than Zafar-2 given any same risk difference.

\section{Related Work}\label{sec:r}



Many methods have been proposed for constructing fairness-aware classifiers, which can be broadly classified into pre/post-processing and in-processing methods. The pre/post-processing methods propose to modify the training data and/or tweak the predictions to obtain fair predictions. Data mining techniques have been proposed to remove bias from a dataset since 2008 \cite{Pedreshi2008}. After that, a number of techniques have been proposed either based on correlations between the sensitive attribute and the decision \cite{Dwork2012,Feldman2015,Wu2016,Zliobaite2011} or the causal relationship among all attributes \cite{Kilbertus2017,Zhang2017,Zhang2017a,Zhang2017b}. In \cite{Hardt2016}, the authors proposed to tweak the output of the classifier after the classifier makes predictions. As suggested by a recent work \cite{zhang2018achieving}, both the pre-processing and post-processing phases are necessary in achieving fair predictions.
Another category of methods are in-processing methods which adjust the learning process of the classifier \cite{Kamishima2011a,agarwal2017reductions,menon2018cost}. In recent years, a number of methods are proposed to incorporate fairness as constraints in the optimization, e.g., \cite{Kamishima2011a,Goh,Krishna2018,Zafar2017b,Zafar2017a,woodworth2017learning,olfat2018spectral}. As discussed in the paper, there lacks a theoretical framework for guiding the formulation of the constrained optimization problem. This paper proposes a general framework for fairness-aware classification.

\section{Conclusions}\label{sec:c}

In this paper, we studied the fairness-aware classification problem and formulated it as the constrained optimization problem. We proposed a general framework which addresses all limitations of previous works in terms of: (1) various fairness metrics can be incorporated into classic classification models as constraints; (2) the formulated constrained optimization problem is convex and can be solved efficiently; and (3) the lower and upper bounds of real fairness measures are established using surrogate functions, which provide a fairness guarantee for our framework. Within the framework, we proposed a constraint-free criterion under which the learned classifier is guaranteed to be fair, as well as developed the method for learning fair classifiers if the constraint-free criterion fails to satisfy. The experimental results using real-world datasets show that our method achieved better fairness performance than previous methods, and also consistently achieved better predictive accuracy under the same fairness performance level. 


\appendix

\section{Proof of Theorem 1}

\begin{proof}
	Following Eq.~(3), the risk difference of the maximum risk difference classifier $f_{\max}$ is given by:
	\begin{eqnarray*}
		\RD(f_{\max}) = \E_{\X} \Big[ \frac{\eta(\x)}{p} \I_{f_{\max}(\X)=1} + \frac{1-\eta(\x)}{1-p} \I_{f_{\max}(\X)=-1} \Big] -1.
	\end{eqnarray*}

	The difference between $\RD ( f_{\max} ) $ and any deterministic classifier $ \RD ( f )$ is given as:
	\begin{align*}
		\RD ( f_{\max} ) - \RD ( f ) & =  \E_{\X} \Big[ \frac{\eta(\x)}{p} \big[\I_{f_{\max}(\X)=1} - \I_{f(\x)=1}\big] \\
		                             & + \frac{1-\eta(\x)}{1-p} \big[\I_{f_{\max}(\X)=-1} - \I_{f(\x)=-1}\big] \Big].
	\end{align*}

	Let's consider the difference of the conditional risk difference:
	\begin{align*}
		DC(\x) & = \frac{\eta(\x)}{p} \big[\I_{f_{\max}(\X)=1} - \I_{f(\x)=1}\big]        \\
		       & + \frac{1-\eta(\x)}{1-p} \big[\I_{f_{\max}(\X)=-1} - \I_{f(\x)=-1}\big],
	\end{align*}
\begin{enumerate}
		\item if $ \eta(\x) \geq p $,  
		\begin{itemize}
			\item if $f(\x) = 1$, 
					$DC(\x)=0$;
			\item if $f(\x) = -1$, 
					$DC(\x)= \frac{\eta(\x)}{p} - \frac{1-\eta(\x)}{1-p} \propto \eta(\x) - p \geq 0 $;
		\end{itemize}
		\item if $ \eta(\x) < p $,  $ f_{\max}(\x) = -1 $, 
		\begin{itemize}
			\item if $f(\x) = 1$, 
					$DC(\x)= -\frac{\eta(\x)}{p} + \frac{1-\eta(\x)}{1-p} \propto -\eta(\x) + p > 0 $;
			\item if $f(\x) = -1$,  
			 		$DC(\x)=0$.
		\end{itemize}
	\end{enumerate}

	We can find the difference of the conditional risk difference $DC(\x)$ is always non-negative. Thus, the difference $\RD ( f_{\max} ) - \RD ( f )$, the weighted average of $DC(\x)$, is also non-negative. So $\RD ( f_{\max} ) \geq \RD ( f ) $ is proved. Similarly, we can readily prove that $ \RD (f_{\min}) \leq \RD ( f ) $.
\end{proof}

\section{Proof of Theorem 3}
\begin{proof}

	Let's firstly verify that $\psi_{\kappa}$ is convex.

	Because $\kappa$ is convex and $ \kappa^{\prime}(0) > 0 $, we have
	\begin{align*}
		H^{\circ}_{\kappa} (\eta) & = \min_{\alpha : \alpha  ( \eta - p ) \geq 0} \frac{\eta}{p}\kappa ( \alpha)  + \frac{1 - \eta}{1-p} \kappa ( - \alpha )                                                                                   \\
		                          & = \min_{\alpha : \alpha  ( \eta - p ) \geq 0}  \big( \frac{\eta }{ p } + \frac{ 1-\eta }{ 1 - p } \big)                                                                                                    \\
		                          & \Big[ \frac{ \frac{\eta}{p} }{ \frac{\eta }{ p } +  \frac{ 1-\eta }{ 1 - p } } \kappa ( \alpha)  + \frac{ \frac{1-\eta}{1-p} }{ \frac{\eta }{ p } + \frac{ 1-\eta }{ 1 - p } } \kappa ( - \alpha )  \Big].
	\end{align*}

	Let $ \nu = \frac{\eta }{ p } + \frac{ 1-\eta }{ 1 - p } $, the above can be reformulated as
	\begin{align*}
		H^{\circ}_{\kappa} (\eta) & = \min_{\alpha : \alpha  ( \eta - p ) \geq 0}  \nu \times \Big[ \frac{\eta}{p \nu } \kappa ( \alpha)  + \frac{1-\eta}{ (1-p) \nu } \kappa ( - \alpha )  \Big].
	\end{align*}

	Since $\kappa$ is convex and according to Jensen's inequality, we can derive
	\begin{align*}
		H^{\circ}_{\phi} (\eta) & \geq \min_{\alpha : \alpha  ( \eta - p ) \geq 0}  \nu \times  \kappa \big( \frac{\eta}{p \nu } \alpha  - \frac{1-\eta}{ (1-p) \nu } \alpha )  \big) \\
		                        & = \min_{\alpha : \alpha  ( \eta - p ) \geq 0}  \nu \times \kappa \big( \frac{ \alpha(\eta - p) }{ \nu * p(1-p) }  \big)   \geq \nu  \kappa (0).
	\end{align*}


	The equality is achieved when $ \alpha  ( \eta - p ) = 0 $, so that
	\[ H^{\circ}_{\kappa} ( \eta ) = \big( \frac{\eta }{ p } + \frac{ 1-\eta }{ 1 - p } \big) \kappa (0) .  \]

	So it follows that
	\[ H^{\circ}_{\kappa} \big( p ( 1- p) \mu + p \big) = ( \mu - 2p\mu + 2) \kappa (0). \]

	Since $H^{\circ}_{\phi}$ and $H^{-}_{\phi}$ are convex ($H^{-}_{\kappa}$ is a point-wise minimum over linear functions and $H^{\circ}_{\kappa}$ is a linear function of $\mu$), we conclude that $ \psi_{\kappa} ( \mu ) = H^{\circ}_{\kappa} ( p ( 1- p) \mu + p ) -  H^{-}_{\kappa} ( p ( 1- p) \mu + p )  $ 	is convex.

	Let's move back to Eq.~(6) whose argument could be rewrite as
	\begin{align*}
		\RD (h) & - \RD^{-}  = \E_\x \big[ \crd (h, \x) \big] - \min_{h \in \mathcal{H}} \E_\x \big[ \crd (h, \x) \big]                            \\
		        & = \E_\x \big[ \crd (h, \x) - \min_{h \in \mathcal{H}} \crd (h, \x) \big]                                                         \\
		        & = \E_\x \Big[ \I_{( \eta - p )  h(\x) < 0} \times \big[ \frac{ \abs{ \eta - p } }{p(1-p)} \big] \Big] = \E_\x \big[ g(\x) \big].
	\end{align*}

	By Jensen's inequality, if $\psi_{\kappa}$ is convex, then we have
	\begin{align*}
		\psi_{\kappa} & \big( \RD(h) - RD^{-} \big)  = \psi_{\kappa} \Big( \E_\x \big[ g(\x) \big] \Big)   \leq  \E_\x \Big[ \psi_{\kappa} \big( g(\x) \big) \Big] \\
		                                           & \leq \E_\x \Big[ \psi_{\kappa} \Big( \I_{( \eta - p )  h(\x) > 0} \big[ \frac{ \abs{ \eta - p } }{p(1-p)} \big]  \Big) \Big] \\
		                                           & =  \E_\x \Big[  \I_{( \eta - p )  h(\x) > 0} \times \psi_{\kappa} \big(  \frac{ \abs{ \eta - p } }{p(1-p)} \big) \Big]       \\
		                                           & =  \E_\x \Big[  \I_{( \eta - p )  h(\x) > 0} \times \big[ H^{\circ}_{\kappa} ( \eta ) - H^{-}_{\kappa} ( \eta ) \big] \Big]  \\
		                                           & = \I_{( \eta - p )  h(\x) > 0} \times \E_\x  \big[ H^{\circ}_{\kappa} ( \eta ) - H^{-}_{\kappa} ( \eta ) \big] .
	\end{align*}

	Note that if $ ( \eta - p )  h(\x) \geq 0 $, we always have $ \crd_{\kappa} \big( h(\x) \big) \geq H^{\circ}_{\kappa} ( \eta ) $ because of the definition of $H^{\circ}_{\kappa}$. Otherwise, we always have $ \crd_{\kappa} \big( h( \x ) \big) \geq H^{-}_{\kappa} ( \eta ) $ because of the definition of $H^{-}_{\phi}$. Thus,
	\begin{align*}
		\psi_{\kappa} & \big( \RD (h)  - \RD^- \big)  \leq \I_{( \eta - p )  h(\x) > 0}                                                                                  \\
		                            & \times \E_\x \big[ H^{\circ}_{\kappa} ( \eta ) - H^{-}_{\kappa} ( \eta ) \big] +  \I_{( \eta - p )  h(\x) \leq 0} \times   0      \\
		                            & \leq \I_{( \eta - p )  h(\x) > 0}  \times \E_\x  \Big[ \crd_{\kappa} \big( h( \x )\big)  - H^{-}_{\kappa} \big( \eta \big) \Big]  \\
		                            & + \I_{( \eta - p )  h(\x) \leq 0} \times   \E_\x  \Big[ \crd_{\kappa} \big( h( \x )\big)  - H^{-}_{\kappa} \big( \eta \big) \Big] \\
		                            & = \E_\x  \Big[ \crd_{\kappa} \big( h( \x )\big)  - H^{-}_{\kappa} \big( \eta \big) \Big]   = \RD_{\kappa} (h) - \RD^{-}_{\kappa}.
	\end{align*}

	Similarly, we can prove $ \psi_{\delta} \big( \RD^+ - \RD (h) \big)  \leq \RD^{+}_{\delta} - \RD_{\delta} (h)$.
	Thus, Theorem~3 is proved.
\end{proof}

\section{Other Fairness Notions}

\textbf{Risk ratio} is a common fairness notion \cite{Pedreschi,Romei2013}. It also requires the decision is independent with the protected attribute. Different with the risk difference, the unfairness is quantified by the ratio of the positive decisions between the non-protected group and the protected group.  Let's formalize the risk ratio $\mathbb{RR}(h)$ of the classifier $h$:
\[ \mathbb{RR} (h) = \frac{ \E_{\X|S=s^+} \big[ \I_{h(\x) > 0}\big] }{ \E_{\X|S=s^-} \big[ \I_{h(\x) > 0}\big] } . \]

The fairness constraints with regards to risk ratio could be expressed as
\[ \mathbb{RR} (h) = \frac{ \E_{\X|S=s^+} \big[ \I_{h(\x) > 0}\big] }{ \E_{\X|S=s^-} \big[ \I_{h(\x) > 0}\big] } \leq \tau. \]

Similar to Eq.~(3), we express the constraints as
\begin{align}
	\label{eq:RR} \E_{\X} \Big[ \frac{\eta}{p} \I_{h(\x) > 0} + \tau \frac{1-\eta(\x)}{1-p} \I_{h(\x) > 0} \Big]  - \tau \leq 0.
\end{align}

\textbf{Equalized odds and equalized opportunity} are proposed by Hardt \etal \cite{Hardt2016}. Equalized odds requires the protected attribute and the predicted label are independent conditional on the truth label. To quantify the strength of equalized odds, we simply propose the prediction difference between two groups conditional on the truth label. So, the equalized odds is
\begin{align*}
	\mathbb{EO} (h) = \E_{\X|S=s^+,Y} [\I_{h(\x)>0}]  - \E_{\X|S=s^-,Y} [\I_{h(\x)>0}].
\end{align*}
Similarly, a classifier $h$ is considered as fair with regard to equalized odds if $ \mathbb{EO} (h) \leq \tau $.

Let's reformulate the equalized odds constraints:
\begin{align}
	\nonumber \mathbb{EO} (h) 
	\nonumber  & = \E_{\X|S=s^+,Y} [\I_{h(\x)>0}]  + \E_{\X|S=s^-,Y} [\I_{h(\x)<0}] - 1     \\
	\nonumber  & = \E_{\X|\Y} \Big[ \frac{P(S=s^+|\x,\y)}{P(S=s^+|\y)} \I_{h(\x)>0}         \\
	\label{EO} & +  \frac{1-P(S=s^+|\x,\y)}{1-P(S=s^+|\y)} \I_{h(\x)<0} \Big] -1 \leq \tau.
\end{align}

Equalized opportunity is a relaxation of equalized odds where only the positive group ( $Y=1$ ) is taken into account: 
\begin{align}
	\nonumber & \mathbb{EOP} (h) = \E_{\X|\Y=1} \Big[ \frac{P(S=s^+|\x,\Y=1)}{P(S=s^+|\Y=1)} \I_{h(\x)>0}      \\
	\label{EOO}                & + \frac{1-P(S=s^+|\x,\Y=1)}{1-P(S=s^+|\Y=1)} \I_{h(\x)<0} \Big] -1 \leq \tau.
\end{align}

By simply replacing the indicator functions with surrogate functions, we can readily extend our framework to the constraints \eqref{eq:RR}, \eqref{EO}, \eqref{EOO} with regard to the three notions. Our criterion and bounds are also extensible to the three notions.


\section*{Acknowledgments}
This work was supported in part by NSF 1646654.

\bibliographystyle{aaai}

\begin{thebibliography}{}

\bibitem[\protect\citeauthoryear{Agarwal \bgroup et al\mbox.\egroup
  }{2017}]{agarwal2017reductions}
Agarwal, A.; Beygelzimer, A.; Dud{\'\i}k, M.; and Langford, J.
\newblock 2017.
\newblock A reductions approach to fair classification.
\newblock In {\em Conference on Fairness, Accountability, and Transparency in
  Machine Learning}.

\bibitem[\protect\citeauthoryear{Bartlett, Jordan, and
  McAuliffe}{2006}]{Bartlett2006}
Bartlett, P.~L.; Jordan, M.~I.; and McAuliffe, J.~D.
\newblock 2006.
\newblock {Convexity, Classification, and Risk Bounds}.
\newblock {\em Journal of the American Statistical Association}
  101(473):138--156.

\bibitem[\protect\citeauthoryear{Diamond and Boyd}{2016}]{StevenDiamond2016}
Diamond, S., and Boyd, S.
\newblock 2016.
\newblock {CVXPY: A Python-Embedded Modeling Language for Convex Optimization}.
\newblock {\em Journal of Machine Learning Research} 17(83):1--5.

\bibitem[\protect\citeauthoryear{Dwork \bgroup et al\mbox.\egroup
  }{2012}]{Dwork2012}
Dwork, C.; Hardt, M.; Pitassi, T.; Reingold, O.; and Zemel, R.
\newblock 2012.
\newblock {Fairness through awareness}.
\newblock In {\em Proceedings of ITCS '12},  214--226.

\bibitem[\protect\citeauthoryear{Feldman \bgroup et al\mbox.\egroup
  }{2015}]{Feldman2015}
Feldman, M.; Friedler, S.~A.; Moeller, J.; Scheidegger, C.; and
  Venkatasubramanian, S.
\newblock 2015.
\newblock {Certifying and Removing Disparate Impact}.
\newblock In {\em KDD '15}.
\newblock ACM Press.

\bibitem[\protect\citeauthoryear{Goh \bgroup et al\mbox.\egroup }{2016}]{Goh}
Goh, G.; Cotter, A.; Gupta, M.; and Friedlander, M.~P.
\newblock 2016.
\newblock Satisfying real-world goals with dataset constraints.
\newblock In {\em NIPS'16},  2415--2423.

\bibitem[\protect\citeauthoryear{Hardt \bgroup et al\mbox.\egroup
  }{2016}]{Hardt2016}
Hardt, M.; Price, E.; Srebro, N.; et~al.
\newblock 2016.
\newblock Equality of opportunity in supervised learning.
\newblock In {\em Advances in neural information processing systems},
  3315--3323.

\bibitem[\protect\citeauthoryear{Kamishima, Akaho, and
  Sakuma}{2011}]{Kamishima2011a}
Kamishima, T.; Akaho, S.; and Sakuma, J.
\newblock 2011.
\newblock {Fairness-aware Learning through Regularization Approach}.
\newblock In {\em ICDMW'11},  643--650.
\newblock IEEE.

\bibitem[\protect\citeauthoryear{Kilbertus \bgroup et al\mbox.\egroup
  }{2017}]{Kilbertus2017}
Kilbertus, N.; Carulla, M.~R.; Parascandolo, G.; Hardt, M.; Janzing, D.; and
  Sch{\"o}lkopf, B.
\newblock 2017.
\newblock Avoiding discrimination through causal reasoning.
\newblock In {\em Advances in Neural Information Processing Systems},
  656--666.

\bibitem[\protect\citeauthoryear{Kleinberg, Mullainathan, and
  Raghavan}{2016}]{Kleinberg2016}
Kleinberg, J.; Mullainathan, S.; and Raghavan, M.
\newblock 2016.
\newblock {Inherent Trade-Offs in the Fair Determination of Risk Scores}.
\newblock  1--23.

\bibitem[\protect\citeauthoryear{Krishna and Williamson}{2018}]{Krishna2018}
Krishna, A., and Williamson, R.~C.
\newblock 2018.
\newblock {The Cost of Fairness in Binary Classification}.
\newblock {\em Proceedings of Machine Learning Research} 81:1--12.

\bibitem[\protect\citeauthoryear{Lichman}{2013}]{Lichman2013}
Lichman, M.
\newblock 2013.
\newblock {UCI Machine Learning Repository}.

\bibitem[\protect\citeauthoryear{Menon and Williamson}{2018}]{menon2018cost}
Menon, A.~K., and Williamson, R.~C.
\newblock 2018.
\newblock The cost of fairness in binary classification.
\newblock In {\em Conference on Fairness, Accountability and Transparency},
  107--118.

\bibitem[\protect\citeauthoryear{Olfat and Aswani}{2018}]{olfat2018spectral}
Olfat, M., and Aswani, A.
\newblock 2018.
\newblock Spectral algorithms for computing fair support vector machines.
\newblock In {\em International Conference on Artificial Intelligence and
  Statistics},  1933--1942.

\bibitem[\protect\citeauthoryear{Pedreschi, Ruggieri, and
  Turini}{2009}]{Pedreschi}
Pedreschi, D.; Ruggieri, S.; and Turini, F.
\newblock 2009.
\newblock {Measuring Discrimination in Socially-Sensitive Decision Records}.
\newblock In {\em Proceedings of the 2009 SIAM International Conference on Data
  Mining}. Philadelphia, PA: Society for Industrial and Applied Mathematics.
\newblock  581--592.

\bibitem[\protect\citeauthoryear{Pedreshi, Ruggieri, and
  Turini}{2008}]{Pedreshi2008}
Pedreshi, D.; Ruggieri, S.; and Turini, F.
\newblock 2008.
\newblock {Discrimination-aware data mining}.
\newblock In {\em KDD 08},  560.
\newblock New York, New York, USA: ACM Press.

\bibitem[\protect\citeauthoryear{Pleiss \bgroup et al\mbox.\egroup
  }{2017}]{Pleiss2017a}
Pleiss, G.; Raghavan, M.; Wu, F.; Kleinberg, J.; and Weinberger, K.~Q.
\newblock 2017.
\newblock On fairness and calibration.
\newblock In {\em NIPS'17}.

\bibitem[\protect\citeauthoryear{Romei and Ruggieri}{2014}]{Romei2013}
Romei, A., and Ruggieri, S.
\newblock 2014.
\newblock {A multidisciplinary survey on discrimination analysis}.
\newblock {\em The Knowledge Engineering Review} 29(05):582--638.

\bibitem[\protect\citeauthoryear{Scott}{2012}]{scott2012}
Scott, C.
\newblock 2012.
\newblock Calibrated asymmetric surrogate losses.
\newblock {\em Electronic Journal of Statistics}  958--992.

\bibitem[\protect\citeauthoryear{Shen \bgroup et al\mbox.\egroup
  }{2016}]{Shen2016}
Shen, X.; Diamond, S.; Gu, Y.; and Boyd, S.
\newblock 2016.
\newblock {Disciplined Convex-Concave Programming}.
\newblock (Cdc):1009--1014.

\bibitem[\protect\citeauthoryear{Woodworth \bgroup et al\mbox.\egroup
  }{2017}]{woodworth2017learning}
Woodworth, B.; Gunasekar, S.; Ohannessian, M.~I.; and Srebro, N.
\newblock 2017.
\newblock Learning non-discriminatory predictors.
\newblock In {\em Conference on Learning Theory},  1920--1953.

\bibitem[\protect\citeauthoryear{Wu and Wu}{2016}]{Wu2016}
Wu, Y., and Wu, X.
\newblock 2016.
\newblock {Using Loglinear Model for Discrimination Discovery and Prevention}.
\newblock In {\em DSAA '16},  110--119.
\newblock IEEE.

\bibitem[\protect\citeauthoryear{Zafar \bgroup et al\mbox.\egroup
  }{2017a}]{Zafar2017b}
Zafar, M.~B.; Valera, I.; {Gomez Rodriguez}, M.; and Gummadi, K.~P.
\newblock 2017a.
\newblock {Fairness Beyond Disparate Treatment {\&} Disparate Impact}.
\newblock In {\em WWW '17},  1171--1180.
\newblock New York, NY, USA: ACM Press.

\bibitem[\protect\citeauthoryear{Zafar \bgroup et al\mbox.\egroup
  }{2017b}]{Zafar2017a}
Zafar, M.~B.; Valera, I.; Rodriguez, M.~G.; and Gummadi, K.~P.
\newblock 2017b.
\newblock {Fairness Constraints: Mechanisms for Fair Classification}.
\newblock In {\em Artificial Intelligence and Statistics}.

\bibitem[\protect\citeauthoryear{Zhang and Bareinboim}{2018}]{Zhang2017}
Zhang, J., and Bareinboim, E.
\newblock 2018.
\newblock Fairness in decision-making â€” the causal explanation formula.
\newblock In {\em AAAI Conference on Artificial Intelligence}.

\bibitem[\protect\citeauthoryear{Zhang and Wu}{2017}]{Zhang2017a}
Zhang, L., and Wu, X.
\newblock 2017.
\newblock {Anti-discrimination learning: a causal modeling-based framework}.
\newblock {\em International Journal of Data Science and Analytics} 4(1):1--16.

\bibitem[\protect\citeauthoryear{Zhang, Wu, and Wu}{2017a}]{Zhang2017c}
Zhang, L.; Wu, Y.; and Wu, X.
\newblock 2017a.
\newblock {A Causal Framework for Discovering and Removing Direct and Indirect
  Discrimination}.
\newblock In {\em IJCAI '17},  3929--3935.

\bibitem[\protect\citeauthoryear{Zhang, Wu, and Wu}{2017b}]{Zhang2017b}
Zhang, L.; Wu, Y.; and Wu, X.
\newblock 2017b.
\newblock {Achieving Non-Discrimination in Data Release}.
\newblock In {\em KDD '17},  1335--1344.
\newblock New York, New York, USA: ACM Press.

\bibitem[\protect\citeauthoryear{Zhang, Wu, and Wu}{2018}]{zhang2018achieving}
Zhang, L.; Wu, Y.; and Wu, X.
\newblock 2018.
\newblock {Achieving Non-Discrimination in Prediction}.
\newblock In {\em IJCAI '18}.

\bibitem[\protect\citeauthoryear{Zliobaite, Kamiran, and
  Calders}{2011}]{Zliobaite2011}
Zliobaite, I.; Kamiran, F.; and Calders, T.
\newblock 2011.
\newblock {Handling Conditional Discrimination}.
\newblock In {\em 2011 IEEE 11th International Conference on Data Mining},
  number~1,  992--1001.
\newblock IEEE.

\end{thebibliography}

\end{document}